\documentclass{article}
\usepackage{spconf,amsmath,graphicx}
\usepackage{comment}
\usepackage{amsmath}
\usepackage{amssymb}
\usepackage{amsfonts}
\usepackage{verbatim}
\usepackage{booktabs}
\usepackage{multirow}
\usepackage{float}
\usepackage{xcolor}
\usepackage{multirow}

\newcommand{\sarimg}{I}
\newcommand{\SarHeight}{H}
\newcommand{\SarWidth}{W}
\newcommand{\SarSpace}{\mathbb{R}^{\SarHeight \times \SarWidth}}

\newcommand{\detectionIndex}{i}
\newcommand{\detectionIndexTwo}{j}

\newcommand{\nDetections}{n}

\newcommand{\frameworkName}{DV\!D}

\newcommand{\nSamplesD}{N}
\newcommand{\nSamplesDOne}{\nSamplesD_1}
\newcommand{\nSamplesDTwo}{\nSamplesD_2}
\newcommand{\nSamplesDThree}{\nSamplesD_3}

\newcommand{\bbox}{b}

\newcommand{\length}{l}
\newcommand{\width}{w}

\newcommand{\class}{\lambda}
\newcommand{\classspace}{\Lambda}

\newcommand{\tonnage}{t}

\newcommand{\dataset}{D}
\newcommand{\dOne}{\dataset_1}
\newcommand{\dTwo}{\dataset_2}
\newcommand{\dThree}{\dataset_3}

\newcommand{\preprocessing}{\mathcal{P}}
\newcommand{\featureExtractor}{\mathcal{F}}
\newcommand{\featureMaps}{F}
\newcommand{\dropoutProbability}{p}

\newcommand{\loss}{\mathcal{L}}
\newcommand{\rpnLoss}{\loss_\text{RPN}}
\newcommand{\detectionLoss}{\loss_\text{det}}
\newcommand{\dimensionsLoss}{\loss_\text{reg}}
\newcommand{\classificationLoss}{\loss_\text{cls}}
\newcommand{\totalLoss}{\loss_\text{total}}
\newcommand{\lossWeight}{\eta}
\newcommand{\lossWeightClass}{\lossWeight_\text{cls}}
\newcommand{\lossWeightReg}{\lossWeight_\text{reg}}

\usepackage{bbding}

\newcommand{\classhead}{\mathcal{C}}
\newcommand{\regrhead}{\mathcal{R}}
\newcommand{\dethead}{\mathcal{D}}
\newcommand{\KNN}{\mathcal{K}}

\newcommand{\shipsPhysical}{(\bbox_\detectionIndex, \length_\detectionIndex, \width_\detectionIndex)}

\newcommand{\predicted}[1]{\widehat{#1}}
\newcommand{\predLength}{\predicted{\length}}
\newcommand{\predWidth}{\predicted{\width}}
\newcommand{\predClass}{\predicted{\class}}
\newcommand{\predTonnage}{\predicted{\tonnage}}
\newcommand{\predBbox}{\predicted{\bbox}}

\newcommand{\knnTotalDistance}{\Delta}
\newcommand{\knnDistance}{\delta}

\newcommand{\knnCatDistance}{\knnDistance_\text{cat}}
\newcommand{\knnCatDistanceParams}{\knnDistance_\text{cat}(\class_\detectionIndex, \class_\detectionIndexTwo)}
\newcommand{\knnLWVector}[1]{\begin{bmatrix}\length_{#1} \\ \width_{#1} \end{bmatrix}}

\newcommand{\knnGaussianRegressor}{f}
\newcommand{\classSpecificRegressor}[1]{f_{#1}}
\newcommand{\wasserstein}{W}

\newcommand{\knnParam}{\alpha}
\newcommand{\knnAgnostic}{\KNN_a}
\newcommand{\knnSpecific}{\KNN_s}
\newcommand{\knnAlpha}{\KNN_\knnParam}

\newcommand{\classOne}{\class_\detectionIndex}
\newcommand{\classTwo}{\class_\detectionIndexTwo}
\newcommand{\knnSurface}{S_{\class_\detectionIndex, \class_\detectionIndexTwo}}
\newcommand{\knnResidual}[1]{r_{#1}}
\newcommand{\knnResidualSet}{R}

\newcommand{\rb}[1]{\textcolor{blue}{Roberto: #1}}

\newcommand{\hOneSpacePre}{\vspace{-2mm}}
\newcommand{\hOneSpacePost}{\vspace{-2mm}}
\newcommand{\hTwoSpacePre}{\vspace{-2mm}}
\newcommand{\hTwoSpacePost}{\vspace{-1mm}}

\newcommand{\eqSpacePre}{\vspace{-2mm}}
\newcommand{\eqSpacePost}{

\vspace{-2mm} \noindent}
\newcommand{\eqShortSpacePost}{

\vspace{-1mm} \noindent}

\newcommand{\parSpacePre}{\vspace{-3mm}}

\newif\ifisReview

\isReviewfalse 

\newcommand{\authorsList}{%
  \ifisReview
    \begin{tabular}{@{}c@{}}
        [Author list redacted for blind review] \\
        \\
    \end{tabular}
  \else
    \begin{tabular}{@{}c@{}}
        Davide Paltrinieri \quad Andrea Diecidue \quad Roberto Basla \quad Daniele Casciani\\
        Piero Fraternali \quad \quad Giacomo Boracchi
    \end{tabular}
  \fi
}

\newcommand{\affiliation}{%
  \ifisReview
    [Affiliation redacted for blind review]%
  \else
    Politecnico di Milano, Milano, Italy
  \fi
}

\usepackage{url}
\usepackage{hyperref}
\newcommand{\github}{%
  \ifisReview
    [redacted for blind review]%
  \else
    \href{https://github.com/PaltrinieriDavide/vesseldetection}{https://github.com/PaltrinieriDavide/vesseldetection}%
  \fi
}

\newcommand{\acknowledgements}{
    \ifisReview
        [Acknowledgements redacted for blind review]%
    \else
        This work was funded by European Union’s Horizon Europe project PERIVALLON - Protecting the EuRopean terrItory from organised enVironmentAl crime through inteLLigent threat detectiON tools, under grant agreement no. 101073952. SAR images were obtained via the Copernicus Browser with assistance from partners in the PERIVALLON project. AIS data has been provided by Kpler through the Kpler MarineTraffic Platform. This work was supported by the Ministry of Education, Youth and Sports of the Czech Republic through the e-INFRA CZ (ID:90254), by the Italian Ministry of University and Research (MUR) under the National Recovery and Resilience Plan (NRRP), and by the European Union (EU) under the NextGenerationEU project.
    \fi
}
\usepackage[font=small,labelfont=bf]{caption}

\title{SAR Vessel Detection and Gross Tonnage Estimation from Heterogeneous Datasets for Dark Vessel Identification}
\name{\authorsList}

\address{\affiliation\vspace{-6mm}}
\begin{document}
%
\maketitle
\begin{abstract}
Detecting vessels engaging in illegal activities is of paramount importance for maritime security. One of the major goals is to detect dark vessels, ships that disable their transponders to evade surveillance. Deep Learning (DL) models can detect vessels in Synthetic Aperture Radar (SAR) images, enabling maritime traffic analysis regardless of weather or visibility conditions. However, to detect potential dark vessels, a DL model must select only those that are required to carry a transponder based on their Gross Tonnage (GT). Unfortunately, no public SAR dataset is available for training an end-to-end DL model for vessel detection and GT regression. In this work, we present a framework that leverages heterogeneous image and tabular datasets to solve this task. Our solution combines a multi-task DL framework for predicting the location, vessel type, and physical dimensions of ships, cascaded with a non-parametric model for predicting GT from vessel size and category. We perform GT regression by a KNN that measures sample similarity using a hybrid Euclidean and categorical distance. Experiments show that our solution can predict multiple outputs while remaining competitive with state-of-the-art models on individual subtasks, thus enabling the identification of dark vessels. We publish our code on GitHub \github.

\end{abstract}
%
\hOneSpacePre
\section{Introduction}
\hOneSpacePost
\label{sec:intro}

Monitoring the maritime environment is crucial for identifying illegal activities at sea. While legitimate ships broadcast their data via the Automatic Identification System (AIS), \textit{dark vessels} purposefully disable their transponders to evade tracking and engage in activities such as smuggling, human trafficking, illegal fishing, and waste disposal~\cite{bueger_beyond_2017}.
Deep Learning (DL) object detection networks for satellite images combined with AIS data analytics represent a promising solution for identifying ships~\cite{hasbi_maritime_2020}. In this work, we focus exclusively on  
Synthetic Aperture Radar (SAR) satellite images as they offer a view of the oceans that is comprehensive, unobstructed and independent from the time of the day, thus enabling the detection of ships having their AIS transponders deactivated.

However, detecting dark vessels from SAR imagery presents significant challenges. First, under international maritime safety conventions, AIS is mandatory only for ships above a GT threshold of 300 on international voyages (500 otherwise), plus all passenger ships~\cite{international_maritime_organization_international_1969}. Therefore, on top of detecting vessels, the network has to predict the vessel class and tonnage to verify whether the transponder of each detected vessel is expected to be active, then search for a match in AIS records for vessel identification. Second, GT measures the overall internal volume of vessels and is difficult to infer from satellite images. This problem is worsened by the low and varying resolution of SAR images across satellites and their noisy nature~\cite{amieva_super-resolution_2024}.
Third, no large, curated and public dataset of SAR images annotated with vessel bounding boxes and GT is available. This prevents training end-to-end DL solutions that detect ships and estimate their GT, as highlighted in Table~\ref{tab:dataset_complementarity}: SSDD~\cite{zhang_sar_2021} and HRSID~\cite{wei_hrsid_2020} provide only bounding box annotations, FUSAR-ship~\cite{hou_fusar-ship_2020} provides only classes and AIS metadata, and xView3-SAR~\cite{paolo_xview3-sar_2022} focuses on dark-fishing activity with coarse classes and length estimates alone. OpenSARShip 2.0~\cite{li_opensarship_2017} is the closest to joint coverage but suffers from scale variability, class imbalance, and incomplete tonnage labels~\cite{toumi_combined_2024}, preventing its application to this task.\footnote{We report in Table \ref{tab:dataset_complementarity} the number of the downloadable records (34528 are declared but only 4391 contain all data needed for our task).}

\begin{figure}[t]
    \centering
    \includegraphics[width=.85\columnwidth]{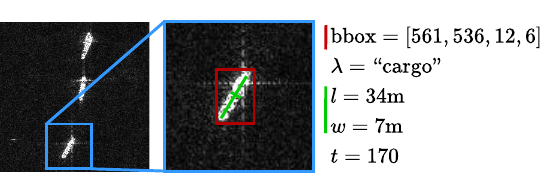}
    \vspace{-6mm}
    \caption{(Left) SAR image from HRSID~\cite{wei_hrsid_2020}. (Right) prediction over the image for the vessel's bounding box $\bbox$, class $\lambda \in \classspace$, length $\length$ and width $\width$ (in meters) and Gross Tonnage $\tonnage$.}
    \label{fig:intro_sar}
    \vspace{-6mm}
\end{figure}

As a result, the literature is fragmented across tasks like detection or classification on task-specific datasets: strong SAR detectors that ignore tonnage~\cite{cao_ship_2024}, multi-task approaches that are trained and tested on private datasets~\cite{dechesne_ship_2019}, or models performing detection and Dead Weight Tonnage (DWT) regression that either rely on optical imagery~\cite{hou_monitoring_2023} or computationally intensive segmentation models and post-processing~\cite{zhang_novel_2025}. Since DWT is a measure of total load capacity as weight instead of volume, these regression methods may transfer poorly to GT regression.

\begin{table}
    \centering
    \caption{Datasets for vessel detection and Gross Tonnage regression (\checkmark: available, --: not available, $\sim$: partially available).}
    \label{tab:dataset_complementarity}
    \scriptsize
    \setlength{\tabcolsep}{1.3pt}
    \vspace{-3mm}
    \begin{tabular}{|l|ccccc|c|c|c|c|}
        \hline
        \multirow{2}{*}{\textbf{Dataset}} & \multicolumn{5}{|c|}{\textbf{Available data}} & \multirow{2}{*}{\textbf{Resolution}} & \multirow{2}{*}{$\mathbf{\SarHeight \times \SarWidth}$} & \multirow{2}{*}{\textbf{\#SAR}} & \multirow{2}{*}{\textbf{\#Vessels}}  \\
        \cline{2-6}
        & SAR & $\bbox$ & $\length$,$\width$ & $\lambda$ & $\tonnage$ &  &  & &  \\
        \hline
        \textbf{OpenSARShip2.0} & \checkmark & \checkmark & \checkmark & \checkmark & $\sim$ & 10 m/px& variable& 46 & 10151\\
        \hline
        \textbf{SSDD} & \checkmark & \checkmark & -- & -- & -- & 1-15 m/px & $668 \times 668$ & 1160 & 2456 \\
        \hline
        \textbf{HRSID} & \checkmark & \checkmark & \checkmark & -- & -- & 0.5--3 m/px & $800 \times 800$ & 5604 & 16951 \\
        \hline
        \textbf{FUSAR} & \checkmark & -- & -- & \checkmark & -- & 1--2 m/px & $512 \times 512$ & 3063 & 3063 \\
        \hline
        \textbf{xView3-SAR} & \checkmark & \checkmark & -- & $\sim$ & - & 20 m/px & $29.4\text{k} \times 24.4\text{k}$  & 991 & 243018 \\
        \hline        \hline
        \textbf{Tab-AIS} & -- & -- & \checkmark & \checkmark & \checkmark & -- & -- & -- & 8096 \\
        \hline
    \end{tabular}
    \vspace{-4mm}
\end{table}

To address these challenges, we propose a solution that combines heterogeneous image and tabular datasets to detect vessels in SAR imagery and estimate their class and Gross Tonnage (Fig. \ref{fig:intro_sar}), thereby enabling automated verification of AIS records to localize dark vessels. To the best of our knowledge, our work is the first to jointly tackle these challenges. We design a custom training procedure for an object detection network across varied image datasets to perform vessel detection, physical dimension regression, and classification. These outputs are fed to a K-Nearest Neighbor regressor based on a hybrid distance measure that fuses a data-driven class distance with the Euclidean distance of the estimated vessel dimensions. This hybrid distance allows us to leverage vessel information from all classes to compensate for the lack of training examples in a single class.
Experimental results show that our framework achieves competitive performance compared with methods focused on either vessel classification or detection. The hybrid distance metric is effective at improving GT predictions for small-support classes.

\hOneSpacePre
\section{Problem Formulation}
\hOneSpacePost
\label{sec:problem_formulation}

Let $\sarimg \in \SarSpace$ be a single-channel input SAR image of height $\SarHeight$ and width $\SarWidth$. The goal of our solution is to map $\sarimg$ to the detection and the physical properties of each of the $\nDetections$ ships that appear in $\sarimg$. In particular, for each ship $\detectionIndex$ we estimate a bounding box $\bbox_\detectionIndex \in \mathbb{R}^4$, the class $\class_\detectionIndex \in \classspace$, and the Gross Tonnage $\tonnage_\detectionIndex \in \mathbb{R}^+$. Our Dark Vessel Detection solution $\frameworkName$ provides a set of detections in the form:
\eqSpacePre
\begin{equation}
    \frameworkName : \sarimg \rightarrow \left\{ \left( \bbox_\detectionIndex, \class_\detectionIndex,  \tonnage_\detectionIndex \right) \right\}_{\detectionIndex=1}^{\nDetections}.
\end{equation}
\eqSpacePost
The peculiarity of our setting is that, as illustrated in Table \ref{tab:dataset_complementarity}, there is no single dataset containing all the annotations for training $\frameworkName$ end-to-end. Thus, our solution is trained on different datasets to address different tasks.



\hOneSpacePre
\section{Methodology}
\begin{figure*}[t]
    \centering
    \includegraphics[width=.95\textwidth]{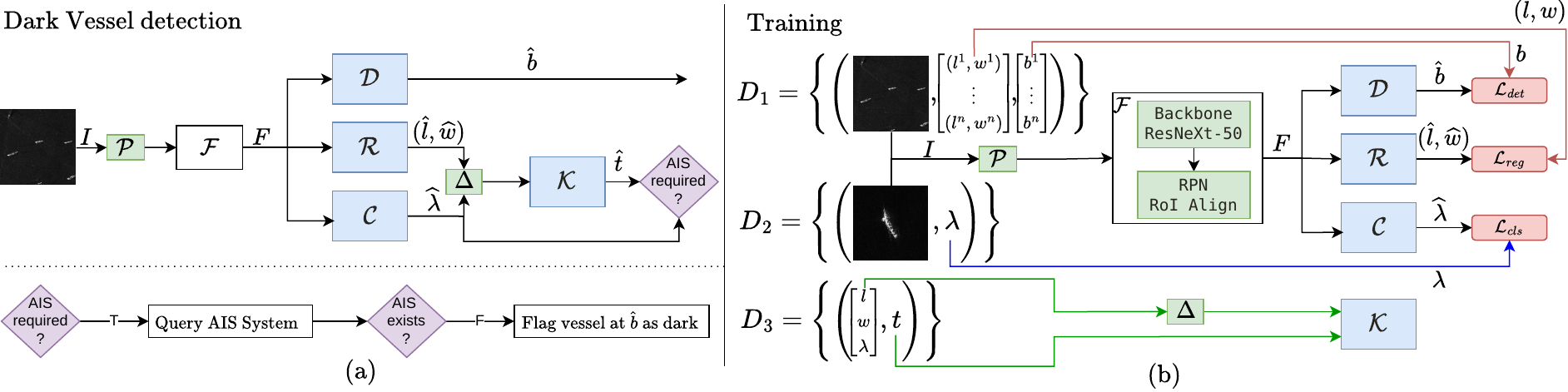}
    \vspace{-4.5mm}
    \caption{Our dark vessel identification solution. (a) Top: A customized object detection network processes a SAR image $\sarimg$ with a preprocessing $\preprocessing$ and a feature extractor $\featureExtractor$. Then, $\dethead$ predicts the bounding box $\predBbox$, while $\regrhead$ and $\classhead$ regress $(\predLength, \predWidth)$ and predict class $\predClass$, respectively. These estimates are fed to a KNN $\KNN$ implementing a hybrid categorical-Euclidean distance $\knnTotalDistance$ to estimate the final Gross Tonnage $\predTonnage$. 
    Bottom: the GT can be used to query the AIS system and, if no vessel is broadcasting the inferred information, flag $\predBbox$ as a dark vessel. (b) We train our network by feeding images $\sarimg$ from SAR datasets $\dOne$ and $\dTwo$. $\featureExtractor$ corresponds to a pretrained ResNeXt-50 backbone, a RPN and a RoI Align stages. During training, features $\featureMaps$ are fed to the three heads, which are trained in phases over the dataset with the corresponding annotation: $\dOne$ optimizes $\dethead$ and $\regrhead$ using the $\detectionLoss$ and $\dimensionsLoss$ losses, while $\dTwo$ trains $\classhead$ with $\classificationLoss$. The categorical distance $\knnTotalDistance$ is fitted on dataset $\dThree$.}
    \label{fig:architecture}
    \vspace{-5.5mm}
\end{figure*}

\hOneSpacePost
\label{sec:method}
Our solution (Fig. \ref{fig:architecture}a) combines first an object detection network based on Faster R-CNN~\cite{ren_faster_2015} with a non-parametric K-Nearest Neighbor model. We customize the DL network with heads for vessel detection ($\dethead$), physical length and width $\predLength, \predWidth \in \mathbb{R}^+$ regression ($\regrhead$), and vessel classification $\predClass \in  \classspace$ ($\classhead$). Then, 
the KNN model takes as input the estimated $\predLength$, $\predWidth$ and $\predClass$ to predict the GT $\predTonnage$ leveraging a hybrid Euclidean-categorical distance. The estimated $\predTonnage$ and $\predClass$ can then be used for querying the AIS system to identify dark vessels as those that are required to broadcast a transponder signal based on $\predTonnage$ but are not found in the records.

\hTwoSpacePre
\subsection{Object Detection Network Architecture}
\hTwoSpacePost
As illustrated in Fig. \ref{fig:architecture}a, we preprocess the input image in the $\preprocessing$ module that \textit{i)} according to~\cite{cao_ship_2024} applies a SAR-specific morphological processing, \textit{ii)} applies Contrast-Limited Adaptive Histogram Equalization (CLAHE)~\cite{ercelik_ship_2025} to improve local contrast and \textit{iii)} linearizes the multiplicative speckle noise through a logarithmic transformation. This preprocessing preserves the resolution of images, enhancing the robustness to resolution shifts coherent with the training set.

In the Faster's feature extraction stage $\featureExtractor$, a ResNext50 backbone extracts the feature maps, which are later processed by the Region Proposal Network (RPN) to generate candidate targets. To accommodate the geometric variability and elongated shapes of vessels, we set the RPN's anchor generator scales in $[8, 256]$ and aspect ratios ranging from $0.2$ to $5.0$. The backbone's output is then processed according to the extracted region proposals with a Multi-Scale RoI Align layer, yielding fixed-size $7 \times 7$ feature representations $\featureMaps$.

Our architecture then branches into the three task-specific heads operating on the feature maps $\featureMaps$ of each vessel as proposed by the RPN proposals. The detection head $\dethead$ refines the proposals and outputs the bounding box coordinates and confidence scores. In parallel, the classification head $\classhead$ uses a dedicated Multi-Layer Perceptron (MLP) to predict the vessel category $\class \in \classspace$. We employ Layer Normalization and a Dropout probability $\dropoutProbability = 0.5$ to mitigate overfitting on the background noise. Finally, the regression head $\regrhead$ is a second MLP designed to estimate the length and width $\predLength$ and $\predWidth$ normalized on the image dimension. 

\hTwoSpacePre
\subsection{Gross Tonnage Estimation} 
\hTwoSpacePost
\label{sec:gt_estimation}
We estimate $\tonnage$ by $\KNN$, a K-Nearest Neighbour (KNN) regressor~\cite{cover_nearest_1967} that takes as input $(\length, \width, \class)$ and predicts the GT $\predTonnage$. First, we fit a regression model for each class $\class$, which we denote as $\knnSpecific$. Given the low dimensional input the regressor can be visualized as a surface for each class (Fig. \ref{fig:knn_curves}a). However, this model might exhibit poor accuracy in regions of the input space where training examples of a specific class are scarce. Thus, we introduce a class-agnostic regressor $\knnAgnostic$ that predicts the Gross Tonnage through interpolation as:
\eqSpacePre
\begin{equation}
    \knnAgnostic\!: \left( \length, \width \right) \rightarrow \predTonnage.
\end{equation}
\eqSpacePost
However, the great variation of tonnage with respect to ship classes (especially at low $\tonnage$) may hinder the prediction. We propose to leverage vessel information from different classes, and define a categorical distance $\knnCatDistance$ between vessels $\detectionIndex$, $\detectionIndexTwo$ of different classes $\class_\detectionIndex$ and $\class_\detectionIndexTwo$. In particular, we define the distance for model $\knnAlpha$ as:
\eqSpacePre
\begin{equation}
    \begin{split}
        \knnTotalDistance \! \left( \detectionIndex, \detectionIndexTwo \right) & = (1 - \alpha) \begin{Vmatrix} \knnLWVector{\detectionIndex} - \knnLWVector{\detectionIndexTwo} \end{Vmatrix}_2 + \alpha \knnCatDistanceParams,
    \end{split}
    \label{eq:cat_distance}
\end{equation}
\eqSpacePost
Where $\knnParam \in [0, 1]$ balances the geometric and categorical distances. Note that $\knnSpecific$ and $\knnAgnostic$  are based on distances that are special cases of (\ref{eq:cat_distance}): in the $\knnAgnostic$ case we have $\knnCatDistanceParams = 0$, while in the $\knnSpecific$ case we have:
\eqSpacePre
\begin{equation}
    \knnCatDistanceParams = \begin{cases}
        0, & \text{if } \class_\detectionIndex = \class_\detectionIndexTwo \\
        \infty, & \text{otherwise}
    \end{cases}.
    \label{eq:delta_cat}
\end{equation}
\eqSpacePost
To leverage samples from different classes, we define two possible ways to compute $\knnCatDistance$.\\ \parSpacePre

\noindent\textbf{Surface distance}
For each vessel class $\class \in \classspace$, we first fit a two-dimensional Gaussian Process Regressor $\knnGaussianRegressor_\class (\length,\width)$ that models tonnage as a smooth function of $\length$ and $\width$. We define the dissimilarity between two vessels' classes $\class_\detectionIndex$ and $\class_\detectionIndexTwo$ as the average absolute difference between two surfaces:
\eqSpacePre
\begin{equation}
    \knnCatDistanceParams = \frac{1}{|\knnSurface|} \iint_{\knnSurface} \!\!\!\!\!\! \left| \knnGaussianRegressor_{\class_\detectionIndex}(\length, \width) - \knnGaussianRegressor_{\class_\detectionIndexTwo}(\length,\width) \right|  d\length  d\width,
\end{equation}
\eqShortSpacePost
Where $\knnSurface$ is the part of the length--width domain where both classes contain training samples. If two classes do not overlap in the $(\length, \width)$ domain, we assign $\knnCatDistanceParams = 1$. Finally, we apply a square-root transformation to the distance matrix and normalize it to the $[0, 1]$ range.\\ \parSpacePre

\noindent\textbf{Residual Distribution Distance} We define $\knnCatDistance$ between two vessels' classes $\classOne$, $\classTwo$ as the prediction error on class $\classOne$ of a model trained on $\classTwo$. Therefore, we train a Gradient Boosting Regressor model \cite{friedman_greedy_2001} $\classSpecificRegressor{\class}: (\length, \width) \rightarrow \predTonnage$ separately for each $\class \in \classspace$. We then compute the residual for each vessel $\detectionIndex$ according to class $\classTwo$ as the difference between its real tonnage and the one predicted by $\classSpecificRegressor{\classTwo}$:
\eqSpacePre
\begin{equation}
    \knnResidual{\classTwo} \! \left( \knnLWVector{\detectionIndex}, \tonnage_\detectionIndex \right) = \tonnage_{\detectionIndex} - \classSpecificRegressor{\classTwo}(\length_\detectionIndex, \width_\detectionIndex), 
\end{equation}
\eqSpacePost
and compute the set of residuals for class $\classOne$ with respect to class $\classTwo$ as:
\eqSpacePre
\begin{equation}
    \knnResidualSet_{\classOne, \classTwo} = \left\{ \knnResidual{\classOne} \! \left( \knnLWVector{\detectionIndexTwo}, \tonnage_\detectionIndexTwo \right)   \right\}.
\end{equation} 
\eqSpacePost
We then define the Residual Distribution Distance function as
\eqSpacePre
\begin{equation}
    \knnCatDistance(\class_\detectionIndex, \class_\detectionIndexTwo) = \frac{\wasserstein(\knnResidualSet_{\classOne, \classOne}, \knnResidualSet_{\classOne, \classTwo})
     + \wasserstein(\knnResidualSet_{\classTwo, \classTwo}, \knnResidualSet_{\classTwo, \classOne})}{2},
\end{equation}
\eqShortSpacePost
Where $\wasserstein(\cdot,\cdot)$ is the Wasserstein distance computed on the residual error distribution on class $\classOne$ predicted by the models for $\classOne$ and $\classTwo$. The result is normalized to the $[0, 1]$ range.

\hTwoSpacePre
\subsection{Datasets}
\hTwoSpacePost

Since no single dataset provides comprehensive SAR image annotations for our task (Sec. \ref{sec:intro}), we leverage three complementary datasets. As described in Fig. \ref{fig:architecture}b, we train the DL model's heads $\dethead$ and $\regrhead$ over HRSID ($\dOne$) as it provides SAR images $\sarimg$ associated to vessel bounding boxes $\bbox$ at known resolution, which allows to compute the physical length $\length$ and width $\width$:
\eqSpacePre
\begin{equation}
    \{ (\sarimg_\detectionIndex, \shipsPhysical) \}_{\detectionIndex=1}^{\nSamplesDOne}.
\end{equation}
\eqSpacePost
We train $\classhead$ using the FUSAR-Ship dataset ($\dTwo$) 
that provides class annotation on images of individual vessels:
\eqSpacePre
\begin{equation}
\{ (\sarimg_{\detectionIndex}, \class_\detectionIndex) \}_{\detectionIndex=1}^{\nSamplesDTwo}.
\end{equation}
\eqShortSpacePost
Finally, we use the tabular dataset Tab-AIS ($\dThree$) that associates a ship's length, width and class to its tonnage to fit the KNN $\KNN$:
\eqSpacePre
\begin{equation}
\{ ((\length_{\detectionIndex}, \width_{\detectionIndex}, \class_\detectionIndex), \tonnage_{\detectionIndex}) \}_{\detectionIndex=1}^{\nSamplesDThree}.
\end{equation}
\eqSpacePost

\hTwoSpacePre
\subsection{Training procedure}
\hTwoSpacePost
To best exploit our fragmented datasets, we train DL components in 3 phases over $\dOne$ and $\dTwo$, and then have a fourth phase for 
$\KNN$ and its categorical distance $\knnCatDistance$ (Fig. \ref{fig:architecture}b). The DL model is optimized with AdamW and a Cosine Annealing LR schedule with specific Early Stopping criteria at each phase. \\ \parSpacePre

\noindent\textbf{Phase 1:} We train the unfrozen backbone initialized with ImageNet weights, the RPN, and heads $\dethead$ and $\regrhead$ on $\dOne$ to predict $\predBbox$, $\predLength$ and $\predWidth$. We use the CIoU detection loss proposed by Zheng et al.~\cite{zheng_distance-iou_2020} as detection loss $\detectionLoss$ for optimizing $\dethead$, while $\regrhead$ is trained with a Smooth L1 loss over the dimensions:
\eqSpacePre
\begin{equation}
    \dimensionsLoss = \text{SmoothL1}(\predLength, \length) + \text{SmoothL1}(\predWidth, \width).
\end{equation}  
\eqSpacePost
We pad each image in the the batch to the maximum dimension and use a learning rate of $10^{-4}$. In this phase, $\classhead$ is frozen and we trigger Early Stopping over the sum of F1 and IoU. \\ \parSpacePre
    
\noindent\textbf{Phase 2:} We train $\classhead$ on $\dTwo$ using Focal Loss~\cite{lin_focal_2017} as $\classificationLoss$ to predict the vessel category. We keep the upper backbone layers (Layer 4 and FPN) unfrozen, while $\regrhead$, $\dethead$, and the RPN remain frozen.
Learning rate is set to $10^{-4}$. We train $\classhead$
with the RPN's region proposals from the feature map $\featureMaps$. Early stopping is based on accuracy (ACC). \\ \parSpacePre
    
\noindent\textbf{Phase 3:} We fine-tune the entire model on datasets $\dOne$ and $\dTwo$. Samples are concatenated and shuffled to be processed in an alternating and randomized mixture of the two datasets. During the forward pass, the model routes data from $\dOne$ to $\dethead$ and $\regrhead$, and data from $\dTwo$ to $\classhead$, computing the losses only over data with the ground truth relevant for each head. In backpropagation, the gradients from the distinct heads are accumulated over multiple steps before performing a single weight update with a lower learning rate ($5 \cdot 10^{-5}$). Thus, 
while the specific heads are updated exclusively by their losses, our combined loss updates the weights of the RPN and backbone. We define this multi-objective loss function as:
\eqSpacePre
\begin{equation}
    \totalLoss = \rpnLoss + \detectionLoss + \lossWeightClass\classificationLoss + \lossWeightReg\dimensionsLoss.
\end{equation} \eqSpacePost \\ \parSpacePre \vspace{-5.5mm}
    
\noindent\textbf{Phase 4:} Finally, we fit the $\KNN$ module on $\dThree$. We first train the $\knnGaussianRegressor_\lambda$ model as defined in Section \ref{sec:gt_estimation} to estimate the categorical distance $\knnCatDistance$. Then, we perform a grid search on the number of neighbors K and the $\knnParam$ parameter with a 5-fold cross-validation to find the optimal KNN hyperparameters.

\begin{figure*}[t]
    \centering
    \includegraphics[width=.92\linewidth]{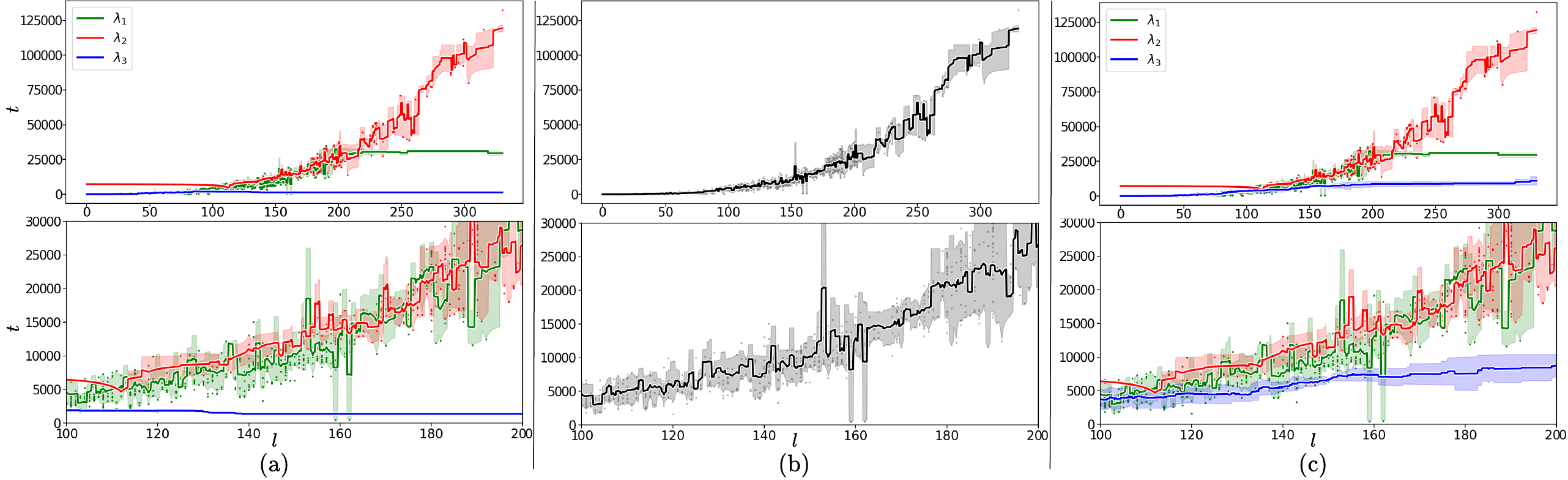}
    \vspace{-4mm}
    \caption{Projected visualization of $\KNN$ regression models on $\dThree$'s training set with $|\classspace| = 3$ in full (top) and detailed (bottom). We project $(\length, \width, \tonnage)$ points and results over the length-tonnage axis by showing predictions for different values of $\width$ in shaded regions (solid lines aggregate the predictions for the same length value). (a) The class-specific model $\knnSpecific$ separates the prediction by classes $\class_1$, $\class_2$ and $\class_3$. (b)  The class agnostic model $\knnAgnostic$ combines all values into a single prediction over all classes. (c) By combining the class-specific models into a $\knnAlpha$ model, different classes can influence the $\predTonnage$ estimation in low-density areas of the $(\length, \width)$ domain for the current class. 
    }
    \label{fig:knn_curves}
    \vspace{-6mm}
\end{figure*}

\hOneSpacePre
\section{Experimental results}
\hOneSpacePost
\label{sec:experiments}

\begin{table}[t]
    \centering
    \scriptsize
    \setlength{\tabcolsep}{4pt}
    \caption{Comparison with related works on SAR images detection, classification, and regression tasks. As baselines, we report the results on classification and detection as in the respective works. We report marked with a (*) values from DWT prediction method reimplemented for our tabular dataset.}
    \vspace{-3mm}
    \label{tab:comparison}
    \begin{tabular}{|l|cc|cc|c|c|}
        \hline
        \multirow{2}{*}{\textbf{Model}} & \multicolumn{4}{|c|}{\textbf{HRSID}}  & \textbf{FUSAR} & \textbf{Tab-AIS} \\
        \cline{2-7}
        & F1 & mAP & MAPE $\length$ & MAPE $\width$ & ACC & MAPE $\tonnage$ \\
        \hline
        Cao et al.      & 89\%   & 91.7\%  & --      & --      & --       & -- \\
        Kamand et al.   & --      & --      & --      & --      & 76.4\% & -- \\
        \hline
        Zhang et al. & --      & 88.5\%      & --      & --      & --       & 32.86\%* \\
        Hou et al.      & --      & --      & --      & --      & --       & 30.41\%* \\
        \hline
        Ours            & 85.7\%  & 85.9\%& 7.57\%  & 11.66\% & 77.3\% & 21.04\% \\
        \hline
    \end{tabular}
    \vspace{-4mm}
\end{table}

\hTwoSpacePre
\subsection{Datasets}
\label{sec:res_datasets}
\hTwoSpacePost
\noindent\textbf{HRSID} The High-Resolution SAR Images Dataset~\cite{wei_hrsid_2020} $\dOne$ comprises images of $800 \times 800$ pixels acquired from Sentinel-1, TerraSAR-X, and TanDEM-X satellites, with resolutions ranging from 0.5 to 3 meters per pixel. HRSID supplies the instance segmentation mask along with the satellite imaging resolution, which allows us to extract the ground truth physical dimensions in pixels by computing the minimum-area rotated bounding box. We convert $\length$ and $\width$ to metric dimensions using the provided per-pixel resolution. We adopt the dataset's train/validation/test splits. \\ \parSpacePre

\noindent\textbf{FUSAR} We use as dataset $\dTwo$ FUSAR-Ship~\cite{hou_fusar-ship_2020}, which provides images of $512 \times 512$ pixels matched with AIS data. To ensure taxonomic alignment with dataset $\dThree$, we restrict our dataset to \textit{general cargo} ($\class_1$, 1,344 vessels), \textit{bulk carriers} ($\class_2$, 170), and \textit{fishing vessels} ($\class_3$, 440),  and split the data in a 70/15/15 split across training, validation and test. \\ \parSpacePre

\noindent\textbf{Tab-AIS} $\dThree$ is a private tabular AIS dataset used for the downstream tonnage estimation stage. Each row provides vessel length, width, category, and Gross Tonnage for $|\classspace| = 6$ vessel classes, reduced to the 3 in common with FUSAR. We split the dataset into training and test with an 80/20 split after outlier and duplicate rows removal. \\ \parSpacePre

\noindent\textbf{Test-SAR} We additionally test our method on a small private dataset combining AIS data and SAR images with resolution 10 m/px collected in the same time frame and location. We devise a semi-automatic labelling procedure for associating the AIS data to vessels in the image by greedily assigning AIS messages to candidate ships extracted by thresholding off-shore areas. According to the distance and time difference between AIS message and image acquisition, we obtain annotations for bounding boxes, dimensions, classes, and GT.  \\ \parSpacePre 

As shown in Table~\ref{tab:dataset_complementarity}, $\dOne$, $\dTwo$ and $\dThree$ are complementary: $\dOne$ provides bounding box and dimensions supervision, $\dTwo$ supplies class categorization, and $\dThree$ contributes the tabular attributes required for the Gross Tonnage estimation.



\begin{table}[t]
    \centering
    \small
    \caption{Performance details of our solutions on the SAR-Test dataset. For $\knnAlpha$, we report the error predicting over the test data $(\length, \width)$ and over the predicted values $(\predLength, \predWidth)$, with higher error in the second case due to error cumulation.}
    \vspace{-3mm}

    \label{tab:sar_test}
    \begin{tabular}{|c|c||c|c||c|}
        \hline
        \multicolumn{2}{|c||}{$\dethead$} & \multicolumn{2}{c||}{$\regrhead$} & $\knnAgnostic$ \\
        \hline
        F1 & mAP & MAPE $\length$ & MAPE $\width$ & MAPE $\tonnage$ \\
        \hline
         79.55\% & 67\% & 18.12\% & 18.45\% & 14.36\% (58.45\%) \\
        \hline
    \end{tabular}
    \vspace{-4mm}
\end{table}

\hTwoSpacePre
\subsection{Results}
\hTwoSpacePost
\label{sec:results} 
Table \ref{tab:comparison} compares our solution against \textit{i)} state-of-the-art single-task baselines evaluated on the same datasets and \textit{ii)} frameworks for the regression of Dead Weight Tonnage (DWT) reimplemented and trained for GT prediction, showing the effectiveness of the proposed method. For FUSAR-Ship classification, we compute the classification accuracy (ACC) and compare it against the dedicated approach by Kamand et al.~\cite{chethan_enhanced_2024}, achieving similar performance. Conversely, Cao et al.~\cite{cao_ship_2024} achieve higher detection metrics on HRSID ($\sim$4\% in F1 score and $\sim$6\% in mean Average Precision, mAP). 
The inferior performance from our model reflects the fact that it has to cope with multiple tasks; nonetheless, the model remains competitive against specialized architectures trained on comparable datasets.

Finally, we evaluate our Gross Tonnage regression against methods by Zhang et al.~\cite{zhang_novel_2025} and Hou et al.~\cite{hou_monitoring_2023}. We re-implemented their linear regression and polynomial equation models as described in the papers to compare the Gross Tonnage prediction against our $\KNN$ module. This substitution yields inaccurate results with up to $11\%$ higher Mean Average Percentage Error (MAPE) with respect to $\KNN$, denoting a low portability of solutions specific for DWT to GT regression.

On our Test-SAR dataset, the model achieves slightly worse performance (Table~\ref{tab:sar_test}), mainly due to the lower resolution of the collected images (Sec. \ref{sec:res_datasets}). We also report the performance of $\knnAgnostic$ when predicting over the datasets $(\length, \width)$ and over the network's outputs $(\predLength, \predWidth)$. MAPE increases drastically in the latter case due to the accumulation of prediction errors at the dimensions regression and GT prediction stages; indeed, as can be seen in Fig. \ref{fig:knn_curves}, an error on $\predLength$ and $\predWidth$ propagates non-linearly to $\predTonnage$.

Our approach averages 0.23s of inference time per $800 \times 800$ image. Measurements have been performed over a set of 1000 images with an average density of 3.13 vessel each. All evaluations have been conducted on a workstation using an Intel Core i9-9900K CPU and an NVIDIA GeForce RTX 2080 Ti GPU (12GB VRAM). \\ \parSpacePre

\noindent\textbf{Ablation of Gross Tonnage regression} We test the impact of our hybrid Euclidean-categorical distances $\knnCatDistance$ (Sec. \ref{sec:gt_estimation}) by showing the GT prediction results on $\dThree$'s test set with all 6 classes. Table \ref{tab:ablation} reports our results, showing that our hybrid distance metric improves upon the baselines especially on low-support classes.


\begin{table}
    \centering
    \small
    \setlength{\tabcolsep}{5pt}
    \caption{Ablation study on $\knnCatDistance$ distances. We evaluate the predicted tonnage $\predTonnage$ on the test set of $\dThree$ with $|\classspace| = 6$  and show that the use of our categorical distances improve the GT regression performance on minority classes. We report aggregated MAPE for this evaluation and the 3 classes used for FUSAR alignment.}
    \label{tab:ablation}
    \vspace{-3mm}
    \footnotesize
    \begin{tabular}{|l|c|c|c|c|c|}
        \hline
        Class & $N$ & $\knnSpecific$ & $\knnAgnostic$ & $\knnAlpha$ (Surf.) & $\knnAlpha$ (Res.) \\ \hline
        Bulk Carrier & 170 & $\mathbf{4.04\%}$ & $4.23\%$ & $4.17\%$ & $4.18\%$ \\ \hline
        Container & 83 & $5.48\%$ & $6.89\%$ & $\mathbf{5.21\%}$ & $\mathbf{5.21\%}$ \\ \hline
        Fishing Vessel & 200 & $27.66\%$ & $\mathbf{24.89\%}$ & $27.78\%$ & $27.93\%$ \\ \hline
        General Cargo & 1252 & $26.58\%$ & $29.11\%$ & $\mathbf{25.33\%}$ & $25.83\%$ \\ \hline
        Oil Tanker & 63 & $16.61\%$ & $19.84\%$ & $16.22\%$ & $\mathbf{15.50\%}$ \\ \hline
        Chem. Tanker & 52 & $10.72\%$ & $7.86\%$ & $7.52\%$ & $\mathbf{5.82\%}$ \\ \hline \hline
        \multicolumn{2}{|c|}{\textbf{MAPE ($|\classspace| = 6$)}} & $22.84\%$ & $24.38\%$ & $\mathbf{21.88\%}$ & $22.17\%$ \\ \hline
        \multicolumn{2}{|c|}{\textbf{MAPE ($|\classspace| = 3$)}} & $21.49\%$ & $22.89\%$ & $21.05\%$ & $\mathbf{21.04\%}$ \\ \hline
    \end{tabular}
    \vspace{-3mm}
\end{table}

\hOneSpacePre
\section{Conclusions}
\hOneSpacePost
In this work, we propose a novel multi-task framework for detecting and identifying dark vessels in SAR imagery. The main challenge we faced was the lack of a dataset for end-to-end training of a DNN solution. We leverage multiple heterogeneous datasets to overcome this issue. Our experimental results demonstrate strong performance compared to state-of-the-art solutions that address only single tasks in our framework. 
While our framework shows strong performance, we have focused on a subset of vessels classified under maritime law
due to a lack of data on most categories; we believe that a solution with more vessel classes that operates in an open-vocabulary setting or based on open-set classification would be more beneficial, especially if employed by law enforcement agencies. Open-vocabulary methods, however, come with much greater hardware demands and higher latency, requiring higher-quality images, stronger feature extractors, and larger training datasets. Open set classification solutions would instead enable the use of class agnostic GT regression models when the detected vessel does not belong to any class. Nonetheless, these are promising future directions for this type of work.

\hOneSpacePre
\section{Acknowledgements}
\hOneSpacePost
\acknowledgements

\bibliographystyle{IEEEbib}
\bibliography{references_2}

\end{document}